\documentclass[conference]{IEEEtran}
\IEEEoverridecommandlockouts

\usepackage{cite}
\usepackage{amsmath,amssymb,amsfonts}
\usepackage{algorithmic}
\usepackage{graphicx}
\usepackage{textcomp}
\usepackage{xcolor}
\usepackage{booktabs}
\usepackage{multirow}
\usepackage{array}
\usepackage{tabularx}
\usepackage{float}
\usepackage{colortbl}
\usepackage{url}
\usepackage[hidelinks]{hyperref}
\usepackage{balance}
\def\BibTeX{{\rm B\kern-.05em{\sc i\kern-.025em b}\kern-.08em
    T\kern-.1667em\lower.7ex\hbox{E}\kern-.125emX}}

\newcommand{\model}{U-PEN Mamba}
\newcommand{\miou}{mIoU}

\newcommand{\R}{\mathbb{R}}

\begin{document}

\title{U-PEN Mamba: Progressive Expansion with Selective State-Space Modeling for Efficient Retinal Vessel Segmentation}

\author{
\IEEEauthorblockN{
Abel A. Reyes-Angulo\textsuperscript{1},
Sidike Paheding\textsuperscript{2},
Vijayan K. Asari\textsuperscript{3},
Mohammad Alam\textsuperscript{4},
and Jeevan Devagiri\textsuperscript{1}
}
\IEEEauthorblockA{
\textsuperscript{1}Michigan Technological University, Houghton, MI, USA \quad
\textsuperscript{2}Fairfield University, Fairfield, CT, USA
}
\IEEEauthorblockA{
\textsuperscript{3}University of Dayton, Dayton, OH, USA \quad
\textsuperscript{4}Minnesota State University, Mankato, MN, USA
}
\IEEEauthorblockA{
areyesan@mtu.edu; spaheding@fairfield.edu; vasari1@udayton.edu; m.alam@mnsu.edu; jdevagir@mtu.edu
}
}

\maketitle

\begin{abstract}
Accurate retinal vessel segmentation is important for computer-aided ophthalmic analysis, yet thin vessels, low contrast, and severe foreground-background imbalance remain challenging for encoder-decoder networks. This paper presents \model, a U-shaped retinal vessel segmentation architecture that couples progressive nonlinear feature expansion with selective state-space modeling. The proposed network enriches local vessel responses with progressive expansion, models long-range spatial dependencies through a Mamba Global Context (MGC) block with linear sequence complexity, and uses attention-based decoder fusion to recover fine vascular boundaries. We evaluate \model\ on CHASE DB1 and DRIVE using a consistent patch-based preprocessing pipeline and compare it with convolutional, attention-based, transformer-based, and Mamba-based segmentation baselines. \model\ obtains the best mean intersection over union among the compared methods, achieving 0.8394 on CHASE DB1 and 0.8221 on DRIVE, with Dice scores of 0.8187 and 0.8078, respectively, using 21.6M trainable parameters. Ablation studies show that the MGC block contributes the largest gain over the U-Net baseline, while projection dimension and state size provide practical accuracy-efficiency control. These results indicate that selective state-space modeling is a promising global-context mechanism for parameter-efficient retinal vessel segmentation. Code is available at: \url{https://github.com/areyesan/UPEN_Mamba}.
\end{abstract}

\begin{IEEEkeywords}
Retinal vessel segmentation, medical image segmentation, U-Net, Mamba, selective state-space models, progressive expansion, deep learning.
\end{IEEEkeywords}

\section{Introduction}
Retinal vessel segmentation is a fundamental step in many ophthalmic image-analysis pipelines. Vessel morphology, branching patterns, and caliber changes are clinically informative for screening and monitoring conditions such as diabetic retinopathy, hypertension, and other systemic or ocular diseases. In practice, however, retinal vessel segmentation is difficult because vessels occupy a small fraction of the image, thin capillaries have low contrast, and imaging artifacts can make vessel boundaries ambiguous.

Deep learning has substantially improved medical image segmentation, especially through convolutional encoder-decoder networks\cite{ramesh2021review,razzak2018deep}. U-Net\cite{ronneberger2015u} remains a strong baseline because it combines hierarchical representation learning with skip connections that recover spatial detail. Its success has motivated many variants, including residual, recurrent, attention-enhanced, nested, and transformer-based models\cite{siddique2021u,diakogiannis2020resunet,alom2018recurrent,oktay2018attention,zhou2018unet++,chen2021transunet}. Despite these advances, the design of efficient segmentation models that preserve fine vascular structure while capturing long-range context is still an open problem.

Attention mechanisms and transformers are effective for global context modeling\cite{vaswani2017attention}, but their computational cost can be high for dense prediction. Recent selective state-space models, particularly Mamba\cite{gu2023mamba}, offer an alternative sequence modeling mechanism with linear complexity in the sequence length: for $L$ spatial tokens, the global-context scan scales as $O(L)$ rather than the $O(L^2)$ pairwise interactions of full self-attention. This property is attractive for medical segmentation, where the network must reason over spatially extended structures without discarding local detail.

Motivated by these considerations, we propose \model, a U-shaped retinal vessel segmentation network that couples local progressive feature expansion with selective state-space global context modeling. Earlier U-PEN variants investigated progressive expansion and attention-based feature selection in U-shaped medical segmentation networks\cite{paheding2022medical,paheding2023u}. The central question addressed here is different: whether a Mamba-style selective state-space module can provide global retinal-vessel context without relying on quadratic self-attention. To answer this, \model\ introduces an MGC block for two-dimensional retinal feature maps and an attention fusion decoder designed to preserve thin-vessel detail.

The main contributions are as follows:
\begin{itemize}
\setlength{\itemsep}{1pt}
\setlength{\parsep}{0pt}
\setlength{\topsep}{2pt}
    \item We propose \model, a U-shaped retinal segmentation network that integrates progressive expansion, residual refinement, selective state-space global context modeling, and attention-based decoder fusion.
    \item We adapt Mamba-style sequence modeling to two-dimensional retinal feature maps through an MGC block that exchanges information across distant vessel regions while retaining linear sequence complexity.
    \item We provide a focused evaluation on CHASE DB1 and DRIVE, showing that \model\ achieves the best \miou\ in comparison against convolutional, attention-based, transformer-based, and U-Mamba baselines.
    \item We analyze component contributions, projection dimension, state size, loss choice, and batch size to clarify where the performance gain comes from and how the model can be scaled.
\end{itemize}

\section{Related Work}
\subsection{CNN-Based Medical Image Segmentation}
Fully convolutional networks\cite{long2015fully} established dense prediction as an end-to-end learning problem. U-Net\cite{ronneberger2015u} extended this idea with an encoder-decoder topology and skip connections, becoming a standard architecture for biomedical segmentation. Subsequent variants introduced residual connections, recurrent refinement, nested skip pathways, and multi-scale feature aggregation\cite{diakogiannis2020resunet,alom2018recurrent,zhou2018unet++}. These designs are effective for local texture and boundary modeling, but the limited receptive field of convolution may still make it difficult to connect thin and spatially separated vessels.

\subsection{Attention and Transformer-Based Segmentation}
Attention modules can suppress irrelevant features and highlight anatomically meaningful regions. Attention U-Net\cite{oktay2018attention} uses gated skip connections to improve feature selection, while transformer-based methods such as TransUNet\cite{chen2021transunet} bring global self-attention into the encoder. These models improve context modeling but often increase memory and computation, particularly when high-resolution spatial tokens are processed by attention.

\subsection{State-Space Models for Vision and Medical Imaging}
Mamba uses input-selective state-space dynamics for sequence modeling\cite{gu2023mamba}. Instead of computing pairwise token interactions as in self-attention, it propagates information through recurrent state updates parameterized by the input. U-Mamba\cite{ma2024u} demonstrated that Mamba-style blocks can be useful for biomedical segmentation. \model\ follows this broader direction but combines selective state-space modeling with progressive expansion and a retinal-vessel-focused decoder fusion strategy.

\subsection{Progressive Expansion in U-Shaped Networks}
U-PEN\cite{paheding2022medical} and U-PEN++\cite{paheding2023u} investigated progressive expansion within U-shaped medical segmentation networks. In this paper, those models are treated as prior work and comparison baselines. The proposed architecture differs by placing selective state-space modeling at the center of the global-context pathway. This design choice is motivated by retinal vessels, where thin structures may extend over long spatial distances and require context beyond local convolution while still preserving high-resolution boundary detail.

\section{Method}
\subsection{Problem Formulation}
Given a color retinal fundus image $\mathbf{X}\in\R^{H\times W\times 3}$, the objective is to estimate a binary vessel mask $\hat{\mathbf{Y}}\in[0,1]^{H\times W}$ that matches the manual annotation $\mathbf{Y}\in\{0,1\}^{H\times W}$. The model is trained on image-mask patches and optimized using a segmentation loss applied to full- and auxiliary-resolution predictions.

\subsection{Architecture Overview}
\begin{figure*}[t]
    \centering
    \includegraphics[width=0.75\textwidth]{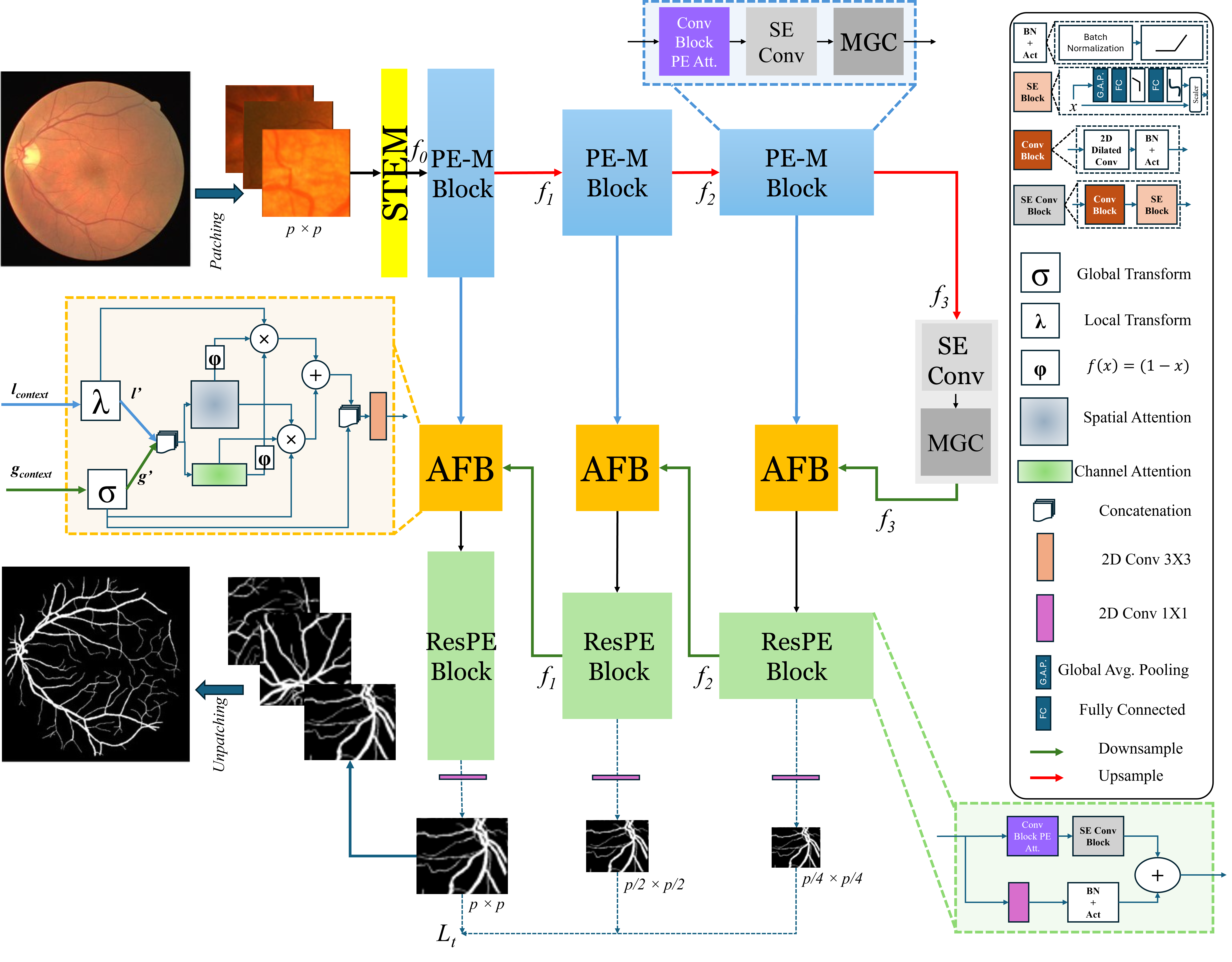}
     \caption{Overview of \model. Progressive expansion enriches local features, the MGC block models long-range context, and AFBs fuse encoder-decoder features for vessel segmentation.}
    \label{fig:upen_mamba}
\end{figure*}

Fig.~\ref{fig:upen_mamba} shows the proposed \model\ architecture. The model follows a U-Net-style encoder-decoder topology with skip connections. The encoder enriches local vessel features through progressive expansion and residual refinement, then applies the MGC block to model long-range spatial dependencies. The decoder upsamples the encoded representation and uses AFBs to combine high-level semantic features with high-resolution encoder details. Thus, global context is handled by selective state-space modeling, while attention is reserved for lightweight decoder fusion.

\subsection{Progressive Expansion Feature Enrichment}
Progressive expansion is used as a local feature-enrichment step before global contextual modeling. For an activation value $x$, a generic $u$-term expansion can be written as
\begin{equation}
    \mathcal{P}_u(x)=\sum_{n=1}^{u} c_n x^{p_n},
    \label{eq:pe}
\end{equation}
where $c_n$ and $p_n$ denote expansion coefficients and powers. In all reported experiments, the block follows the PEL-3 setting, i.e., a third-order expansion with powers $p_n\in\{1,2,3\}$; the following learnable convolutions reweight the expanded responses, so unhelpful high-order terms can be suppressed during training. Applied channel-wise, the expansion is used as local feature-basis augmentation rather than as a replacement for convolution or dilation.

\subsection{Mamba Global Context Block}
The MGC block adapts selective state-space modeling to two-dimensional feature maps. Given an intermediate tensor $\mathbf{F}\in\R^{B\times C\times H\times W}$, the spatial dimensions are flattened into a sequence $\mathbf{Z}\in\R^{B\times L\times C}$, where $L=HW$. A linear projection maps the sequence to the internal Mamba dimension, and the selective state-space operation propagates information along the spatial sequence. A simplified form of the state update is
\begin{align}
    \mathbf{h}_{k} &= \bar{\mathbf{A}}_{k}\mathbf{h}_{k-1}+\bar{\mathbf{B}}_{k}\mathbf{z}_{k}, \\
    \mathbf{o}_{k} &= \mathbf{C}_{k}\mathbf{h}_{k},
    \label{eq:ssm}
\end{align}
where $\mathbf{z}_{k}$ is the $k$-th spatial token, $\mathbf{h}_{k}$ is the hidden state, and the state-space parameters are input dependent. Because the scan updates each token once, the dominant sequence operation is linear in $L=HW$. The output sequence is normalized, projected back to the original channel dimension, reshaped to $B\times C\times H\times W$, and added through a residual connection:
\begin{equation}
    \mathbf{F}_{\mathrm{mgc}} = \mathbf{F} + \operatorname{Reshape}\left(\operatorname{Proj}\left(\operatorname{Mamba}(\mathbf{Z})\right)\right).
    \label{eq:mgc}
\end{equation}
This block allows the encoder to exchange information across distant vessel regions without quadratic self-attention over all spatial locations.

\subsection{Residual Progressive Expansion Block}
The residual progressive expansion block stabilizes nonlinear enrichment by combining an expansion branch with a shortcut branch. For an input feature tensor $\mathbf{F}$, the expansion branch applies progressive expansion, convolution, normalization, activation, and squeeze-excitation recalibration. The shortcut branch aligns dimensions with a $1\times1$ convolution when needed. The block output is
\begin{equation}
    \mathbf{Y}=\operatorname{SE}\left(\operatorname{Conv}_{3\times3}\left(\mathcal{P}_{u}(\mathbf{F})\right)\right)+
    \operatorname{BN}\left(\operatorname{Conv}_{1\times1}(\mathbf{F})\right).
    \label{eq:respe}
\end{equation}
The residual path improves gradient flow and helps preserve fine vessel cues while the expansion branch increases local feature diversity.

\subsection{Attention Fusion Decoder}
Skip connections are essential for retaining thin vessel boundaries, but direct concatenation can also pass irrelevant texture. The AFB adaptively combines local encoder features and global decoder features. Let $\mathbf{G}$ be a high-level decoder feature map and $\mathbf{L}$ be the aligned local feature map from the encoder. After $1\times1$ projections,
\begin{equation}
    \tilde{\mathbf{G}}=\operatorname{Conv}_{1\times1}(\mathbf{G}), \quad
    \tilde{\mathbf{L}}=\operatorname{Conv}_{1\times1}(\mathbf{L}).
\end{equation}
The concatenated tensor is used to compute spatial and channel gates:
\begin{align}
    \mathbf{A}_{s} &= \sigma\left(\operatorname{Conv}_{3\times3}\left([\tilde{\mathbf{G}},\tilde{\mathbf{L}}]\right)\right), \\
    \mathbf{A}_{c} &= \sigma\left(\operatorname{Conv}_{1\times1}\left(\operatorname{ReLU}\left(\operatorname{BN}\left(\operatorname{Conv}_{1\times1}\left([\tilde{\mathbf{G}},\tilde{\mathbf{L}}]\right)\right)\right)\right)\right),
\end{align}
where $\sigma$ denotes sigmoid activation, $[\cdot,\cdot]$ denotes channel concatenation, and $\odot$ denotes element-wise multiplication. The fused feature is
\begin{equation}
    \mathbf{F}_{\mathrm{afb}} = \tilde{\mathbf{G}}\odot\mathbf{A}_{s}\odot\mathbf{A}_{c}
    + \tilde{\mathbf{L}}\odot(1-\mathbf{A}_{s})\odot(1-\mathbf{A}_{c}),
\end{equation}
followed by a $3\times3$ convolution. This design lets the decoder retain local vessel boundaries while using global features to reduce false positives in background regions.

\subsection{Multi-Scale Supervision}
\model\ produces one full-resolution prediction and two auxiliary predictions. Let $\ell(\hat{\mathbf{Y}},\mathbf{Y})$ denote the segmentation loss. The total objective is
\begin{equation}
\mathcal{L}=\phi_1\ell(\hat{\mathbf{Y}}_{1},\mathbf{Y}_{1})+
\phi_2\ell(\hat{\mathbf{Y}}_{1/2},\mathbf{Y}_{1/2})+
\phi_3\ell(\hat{\mathbf{Y}}_{1/4},\mathbf{Y}_{1/4}),
\label{eq:loss}
\end{equation}
where the weights are $\phi_1=1.0$, $\phi_2=0.5$, and $\phi_3=0.25$. The lower-resolution targets are obtained by resizing the ground-truth mask to the corresponding output scale. Unless otherwise stated, the experiments use Dice-BCE loss.

\section{Experimental Setup}
\subsection{Datasets}
We evaluate on CHASE DB1\cite{chase} and DRIVE\cite{drive}, two public benchmarks for retinal vessel segmentation. CHASE DB1 contains retinal fundus images with manual vessel annotations, and DRIVE provides a standard split for digital retinal image vessel extraction. These datasets differ in acquisition conditions and vessel appearance, making them useful for testing robustness across retinal imaging settings.


\subsection{Preprocessing and Augmentation}
Images are converted to RGB and masks to binary labels. Full images are split at the image level before patch extraction to avoid patch-level leakage. Each image is divided into $192\times192$ patches, and 75 patches are sampled per image with a fixed random seed. Training patches are augmented with $90^{\circ}$, $180^{\circ}$, and $270^{\circ}$ rotations and horizontal flipping. Metrics are computed on patch predictions under this protocol; full-image stitching is left for future deployment-oriented evaluation.

\subsection{Training Protocol and Metrics}
The main \model\ configuration uses filters $[64,128,256,512]$, projection dimension 256, 8 Mamba states, dropout 0.2, dilation 2 in the dilated convolutional branches, and squeeze-excitation reduction 16. The model is trained for 60 epochs with AdamW, learning rate $10^{-4}$, weight decay $10^{-5}$, ReduceLROnPlateau scheduling, mixed precision, and batch size 8. Performance is reported using Dice coefficient and mean intersection over union (\miou). Tables~\ref{tab:chase_results} and~\ref{tab:drive_results} combine baselines evaluated in the same patch protocol with retinal-specific literature-only rows; rows with missing \miou\ are included for context and are not protocol-matched. For a binary prediction $\hat{Y}$ and ground truth $Y$, these metrics are
\begin{align}
    \operatorname{Dice}(\hat{Y},Y) &= \frac{2|\hat{Y}\cap Y|}{|\hat{Y}|+|Y|}, \\
    \operatorname{IoU}(\hat{Y},Y) &= \frac{|\hat{Y}\cap Y|}{|\hat{Y}\cup Y|}.
\end{align}

\section{Results}
\subsection{Quantitative Comparison}
Tables~\ref{tab:chase_results} and~\ref{tab:drive_results} compare \model\ with convolutional, attention-based, transformer-based, Mamba-based, and retinal-specific baselines. On CHASE DB1, \model\ obtains the best \miou\ and Dice among the compared methods. On DRIVE, it obtains the best \miou\ and the second-best Dice. Since some retinal-specific methods report Dice but not \miou, the main conclusion is that \model\ provides the strongest reported overlap performance among methods with available IoU values on both datasets.


\begin{table}[t]
\centering
\caption{Segmentation performance on CHASE DB1. Best values are shown in bold and second-best values are underlined. Missing values indicate metrics not reported.}
\label{tab:chase_results}
\scriptsize
\setlength{\tabcolsep}{2pt}
\renewcommand{\arraystretch}{0.95}
\begin{tabularx}{\columnwidth}{@{}
>{\raggedright\arraybackslash}X
>{\centering\arraybackslash}p{0.46in}
>{\centering\arraybackslash}p{0.43in}
>{\centering\arraybackslash}p{0.43in}
@{}}
\toprule
\textbf{Method} &
\textbf{Params} &
\textbf{mIoU} &
\textbf{Dice} \\
&
\textbf{(M)} &
&
\\
\midrule
U-Net\cite{ronneberger2015u} & 34.5 & 0.7567 & 0.7033 \\
ResUNet\cite{diakogiannis2020resunet} & 18.7 & 0.7762 & 0.7922 \\
Att. U-Net\cite{oktay2018attention} & 10.2 & 0.8131 & 0.7980 \\
R2U-Net\cite{alom2018recurrent} & 23.0 & 0.8119 & 0.7960 \\
$U^2$-Net\cite{qin2020u2} & 52.9 & 0.8116 & 0.7961 \\
U-Net++\cite{zhou2018unet++} & 34.5 & 0.8001 & 0.7810 \\
TransUNet\cite{chen2021transunet} & 199.0 & 0.8075 & 0.7900 \\
U-KAN\cite{li2024u} & 6.3 & 0.8131 & 0.7980 \\
U-Mamba\cite{ma2024u} & 47.8 & 0.8004 & 0.7785 \\
Iternet\cite{li2020iternet} & -- & -- & 0.8073 \\
Liu et al.\cite{liu2023retina} & 15.6 & -- & \underline{0.8162} \\
\midrule
U-PEN\cite{paheding2022medical} & 19.4 & 0.7789 & 0.7929 \\
U-PEN++\cite{paheding2023u} & 4.5 & 0.8098 & 0.8062 \\
\textbf{\model} & 21.6 & \textbf{0.8394} & \textbf{0.8187} \\
\bottomrule
\end{tabularx}
\end{table}

\begin{table}[t]
\centering
\caption{Segmentation performance on DRIVE. Best values are shown in bold and second-best values are underlined. Missing values indicate metrics not reported.}
\label{tab:drive_results}
\scriptsize
\setlength{\tabcolsep}{2pt}
\renewcommand{\arraystretch}{0.95}
\begin{tabularx}{\columnwidth}{@{}
>{\raggedright\arraybackslash}X
>{\centering\arraybackslash}p{0.46in}
>{\centering\arraybackslash}p{0.43in}
>{\centering\arraybackslash}p{0.43in}
@{}}
\toprule
\textbf{Method} &
\textbf{Params} &
\textbf{mIoU} &
\textbf{Dice} \\
&
\textbf{(M)} &
&
\\
\midrule
U-Net\cite{ronneberger2015u} & 34.5 & 0.7784 & 0.6928 \\
ResUNet\cite{diakogiannis2020resunet} & 18.7 & 0.7908 & 0.7872 \\
Att. U-Net\cite{oktay2018attention} & 10.2 & 0.8050 & 0.7971 \\
R2U-Net\cite{alom2018recurrent} & 23.0 & 0.7979 & 0.7895 \\
$U^2$-Net\cite{qin2020u2} & 52.9 & 0.7946 & 0.7852 \\
U-Net++\cite{zhou2018unet++} & 34.5 & 0.7765 & 0.7603 \\
TransUNet\cite{chen2021transunet} & 199.0 & \underline{0.8083} & 0.8002 \\
U-KAN\cite{li2024u} & 6.3 & 0.7781 & 0.7840 \\
U-Mamba\cite{ma2024u} & 47.8 & 0.7937 & 0.7833 \\
Li et al.\cite{li2022dual} & 14.1 & -- & 0.7994 \\
Liu et al.\cite{liu2023retina} & 15.6 & -- & \textbf{0.8291} \\
\midrule
U-PEN\cite{paheding2022medical} & 19.4 & 0.7910 & 0.7852 \\
U-PEN++\cite{paheding2023u} & 4.5 & 0.8007 & 0.7909 \\
\textbf{\model} & 21.6 & \textbf{0.8221} & \underline{0.8078} \\
\bottomrule
\end{tabularx}
\end{table}

\subsection{Qualitative Results}
\begin{figure}[t]
    \centering
    \includegraphics[width=0.44\textwidth]{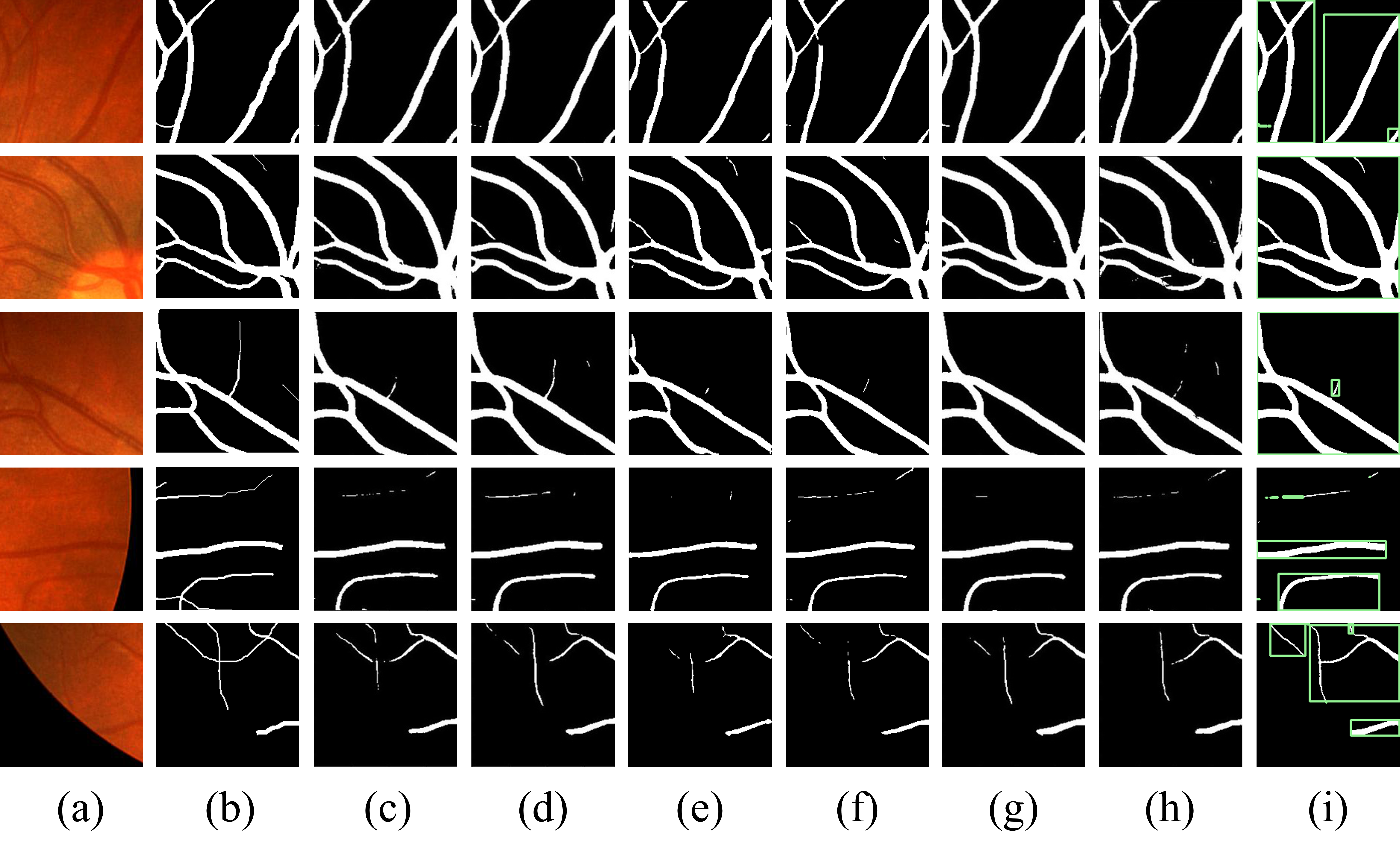}
    \caption{CHASE DB1 qualitative comparison: (a) input, (b) ground truth, (c)--(h) baselines, and (i) \model.}
    \label{fig:results_chase}
\end{figure}

\begin{figure}[t]
    \centering
    \includegraphics[width=0.44\textwidth]{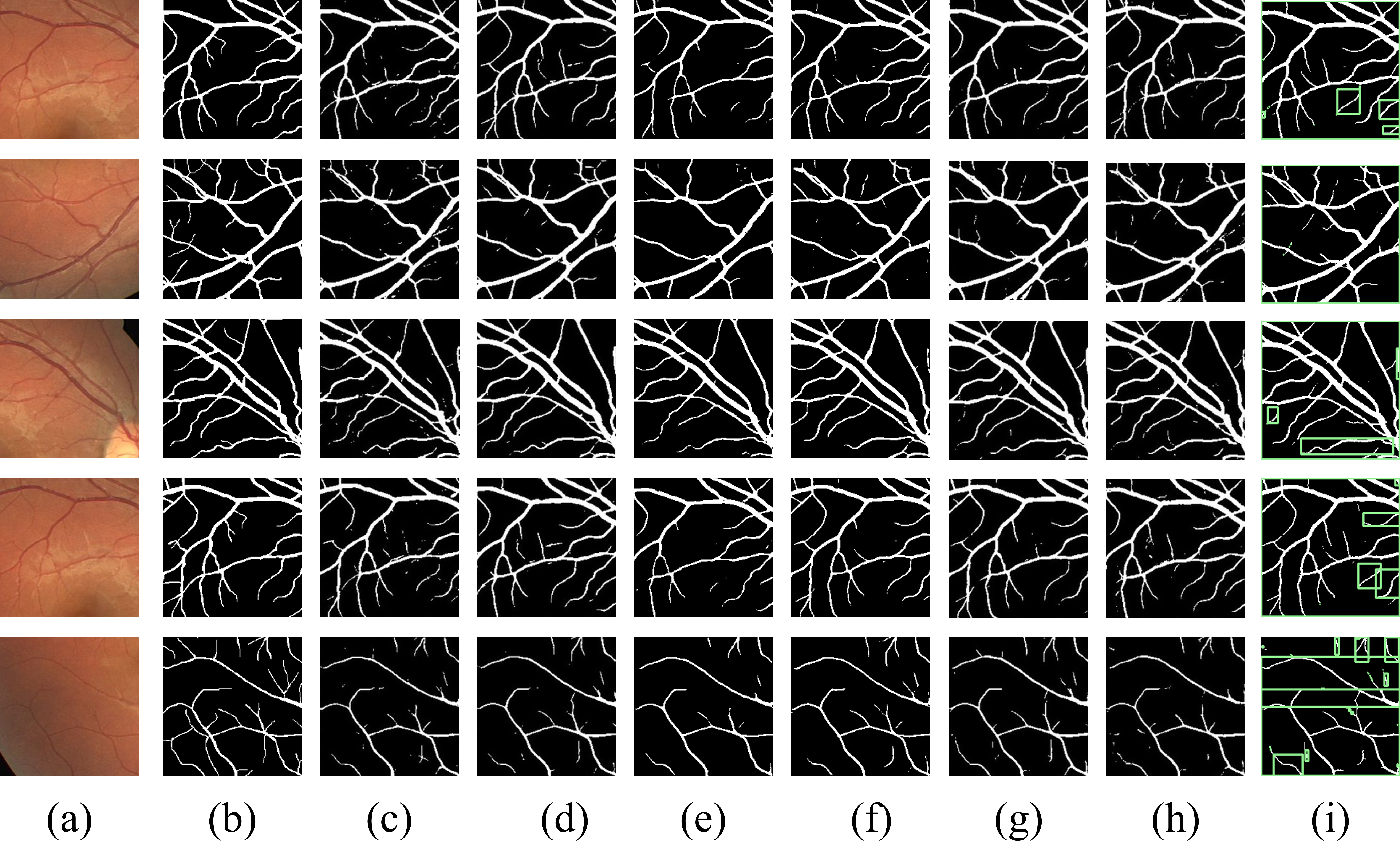}
    \caption{DRIVE qualitative comparison: (a) input, (b) ground truth, (c)--(h) baselines, and (i) \model.}
    \label{fig:results_drive}
\end{figure}

Figs.~\ref{fig:results_chase} and~\ref{fig:results_drive} show qualitative comparisons. The proposed model preserves more continuous vessel traces in difficult regions and reduces background confusion compared with conventional U-shaped baselines. These trends are consistent with the \miou\ gains and suggest improved contextual continuity.

\subsection{Accuracy-Efficiency Behavior}
\begin{figure}[t]
    \centering
    \includegraphics[width=0.44\textwidth]{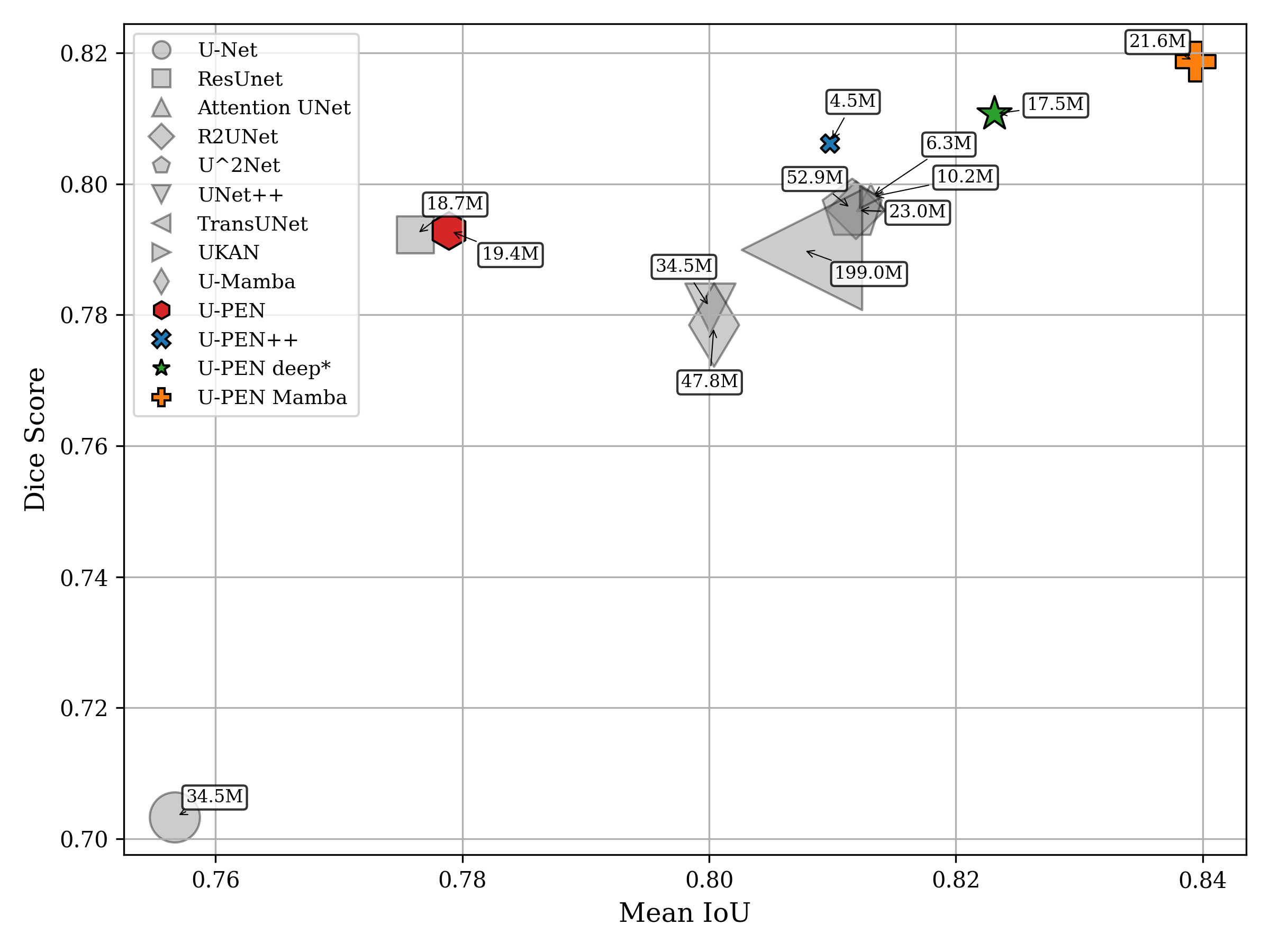}
    \caption{Parameter count versus segmentation performance on CHASE DB1.}
    \label{fig:parameters_chase}
\end{figure}

\begin{figure}[t]
    \centering
    \includegraphics[width=0.44\textwidth]{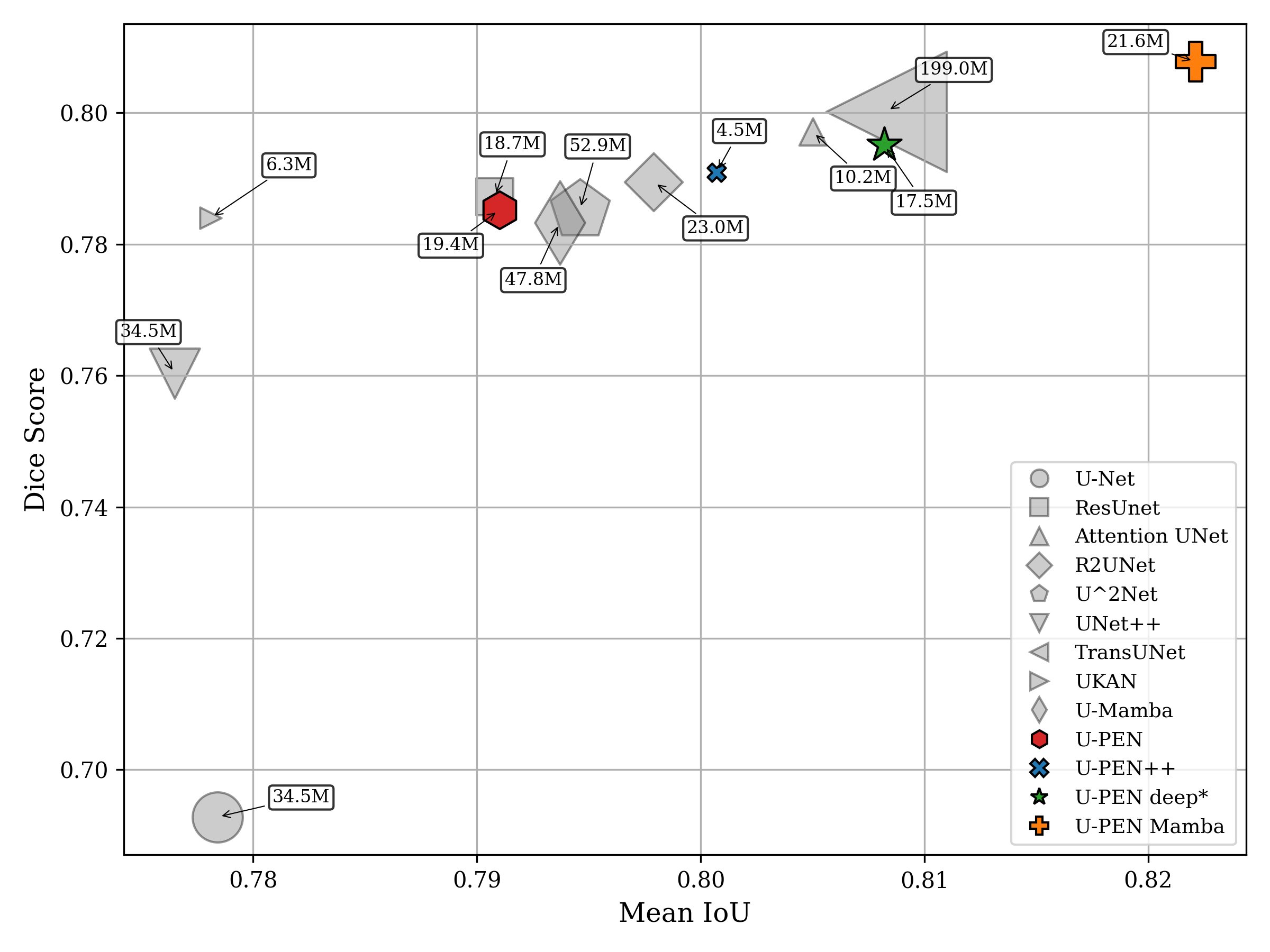}
    \caption{Parameter count versus segmentation performance on DRIVE.}
    \label{fig:parameters_drive}
\end{figure}

Fig.~\ref{fig:parameters_chase} and Fig.~\ref{fig:parameters_drive} summarize model size versus segmentation quality. \model\ uses fewer parameters than several large baselines, including U-Net, TransUNet, and U-Mamba, while achieving the highest \miou\ in both datasets. Since FLOPs, latency, and memory were not measured for every baseline, the efficiency claim is restricted here to parameter efficiency.

\section{Ablation and Scalability Analysis}
\subsection{Component Ablation}
Table~\ref{tab:ablation} summarizes the component ablation on CHASE DB1. Adding progressive expansion improves both overlap metrics relative to the U-Net baseline, and adding the MGC block produces the largest additional gain. The DRIVE sensitivity trends in Tables~\ref{tab:proj_dim}--\ref{tab:batch} are consistent with the selected final configuration, but a full DRIVE component ablation is left for extended validation.

\begin{table}[H]
\centering
\caption{Component ablation on CHASE DB1.}
\label{tab:ablation}
\scriptsize
\begin{tabular}{ccc|cc}
\toprule
\textbf{Base U-Net} & \textbf{+ PE} & \textbf{+ MGC} & \textbf{Mean IoU} & \textbf{Dice} \\
\midrule
\checkmark & -- & -- & 0.7567 & 0.7033 \\
\checkmark & \checkmark & -- & 0.7789 & 0.7929 \\
\checkmark & \checkmark & \checkmark & \textbf{0.8394} & \textbf{0.8187} \\
\bottomrule
\end{tabular}
\end{table}

\subsection{Projection Dimension and State Size}
Tables~\ref{tab:proj_dim} and~\ref{tab:states} show how projection dimension and the number of Mamba states influence Dice score and parameter count. Increasing the projection dimension from 32 to 256 slightly increases model size and gives the best Dice score in the reported configurations. Increasing the number of states beyond 8 does not consistently improve accuracy, suggesting that this retinal vessel setting benefits more from an appropriate projection dimension than from a larger state dimension.

\begin{table}[H]
\centering
\caption{Effect of projection dimension with 8 Mamba states.}
\label{tab:proj_dim}
\scriptsize
\begin{tabular}{cccc}
\toprule
\textbf{Projection dim.} & \textbf{Params} & \textbf{CHASE Dice} & \textbf{DRIVE Dice} \\
\midrule
32  & 20,638,262 & 0.8161 & 0.8052 \\
64  & 20,718,614 & 0.8179 & 0.8061 \\
128 & 20,936,534 & 0.8172 & 0.8044 \\
256 & 21,651,030 & \textbf{0.8187} & \textbf{0.8078} \\
\bottomrule
\end{tabular}
\end{table}

\begin{table}[H]
\centering
\caption{Effect of the number of Mamba states at projection dimension 256.}
\label{tab:states}
\scriptsize
\begin{tabular}{cccc}
\toprule
\textbf{States} & \textbf{Params} & \textbf{CHASE Dice} & \textbf{DRIVE Dice} \\
\midrule
8  & 21,651,030 & 0.8187 & \textbf{0.8078} \\
16 & 23,656,662 & \textbf{0.8189} & 0.8036 \\
32 & 23,803,094 & 0.8177 & 0.8057 \\
\bottomrule
\end{tabular}
\end{table}

\subsection{Loss Function and Batch Size}
Tables~\ref{tab:loss} and~\ref{tab:batch} report additional training sensitivity analyses. Dice-BCE gives the strongest performance among the tested losses, which is consistent with the class-imbalanced nature of binary vessel segmentation. Batch size 8 provides the best Dice score on both datasets, while larger batches slightly reduce performance in this training setup.

\begin{table}[H]
\centering
\caption{Dice coefficient under different loss functions.}
\label{tab:loss}
\scriptsize
\begin{tabular}{lcc}
\toprule
\textbf{Loss} & \textbf{CHASE DB1} & \textbf{DRIVE} \\
\midrule
Dice-BCE & \textbf{0.8187} & \textbf{0.8078} \\
Lovasz & 0.7671 & 0.7937 \\
Focal-Tversky & 0.8016 & 0.7770 \\
Contour-aware & 0.7981 & 0.7923 \\
\bottomrule
\end{tabular}
\end{table}

\begin{table}[H]
\centering
\caption{Dice coefficient under different batch sizes.}
\label{tab:batch}
\scriptsize
\begin{tabular}{ccc}
\toprule
\textbf{Batch size} & \textbf{CHASE DB1} & \textbf{DRIVE} \\
\midrule
4  & 0.8166 & 0.8046 \\
8  & \textbf{0.8187} & \textbf{0.8078} \\
16 & 0.8169 & 0.8034 \\
32 & 0.8136 & 0.8048 \\
\bottomrule
\end{tabular}
\end{table}

\section{Discussion}
The results indicate that \model\ is strongest in \miou, which measures overlap between predicted and annotated vessel regions. This is important for retinal vessel segmentation because broken thin vessels and small boundary errors can substantially affect overlap-based evaluation. The MGC block helps maintain vessel continuity across spatially separated regions, while the AFB decoder preserves local boundary detail.

Compared with transformer-based segmentation, \model\ avoids quadratic self-attention in the global-context module and instead uses selective state-space modeling with linear sequence complexity. Compared with U-Mamba, it uses fewer parameters and obtains higher \miou\ on both retinal benchmarks. Relative to earlier U-PEN variants~\cite{paheding2022medical,paheding2023u}, \model\ is not the smallest model, but it provides the strongest overlap accuracy among the U-PEN-family comparison rows. Thus, it is most suitable when segmentation quality is prioritized under a moderate parameter budget.

The study has limitations. The experiments use two public retinal datasets; broader validation on additional datasets, devices, and cross-dataset settings is needed. The reported results are single-run aggregate values, so paired statistical testing and confidence intervals remain future work. The efficiency analysis is based on parameter count and asymptotic complexity; FLOPs, latency, memory usage, and full-image stitching analysis should be reported in future experiments. Finally, the current ablations isolate PE and MGC, but do not separately evaluate AFB, SE, multi-scale supervision, or alternative global-context modules.

\section{Conclusion}
This paper presented \model, a retinal vessel segmentation network that combines progressive feature expansion with selective state-space global context modeling. The architecture uses residual progressive expansion for local nonlinear enrichment, an MGC block for long-range spatial dependency modeling, and attention fusion for decoder refinement. On CHASE DB1 and DRIVE, \model\ achieves the best \miou\ among the compared methods and competitive Dice scores against specialized retinal segmentation methods. Ablation and scalability analyses show that MGC is the main contributor to the performance gain, while projection dimension and state size provide practical accuracy-efficiency control. These results support selective state-space modeling as a promising direction for accurate retinal vessel segmentation.

\begingroup
\scriptsize
\balance
\bibliographystyle{IEEEtran}
\bibliography{UPEN_Mamba_ICMLA_26_post_acceptance_refs}
\endgroup

\end{document}